\def\year{2021}\relax
\documentclass[letterpaper]{article} 
\usepackage{aaai21}  
\usepackage{times}  
\usepackage{helvet} 
\usepackage{courier}  
\usepackage[hyphens]{url}  
\usepackage{graphicx} 
\usepackage{natbib}  
\usepackage{caption} 
\title{DASZL: Dynamic Action Signatures for Zero-shot Learning}

\author{
    Tae Soo Kim\textsuperscript{*},
    Jonathan Jones\textsuperscript{*} ,
    Michael Peven\thanks{Authors contributed equally to this work.},\\ 
    Zihao Xiao  ,
    Jin Bai  ,
    Yi Zhang  ,
    Weichao Qiu ,
    Alan Yuille  ,
    Gregory D. Hager \\
    
}
\affiliations{
    Johns Hopkins University\\

    3400 N. Charles Street\\
    Baltimore, Maryland 21218\\
    \{tkim60, jdjones , mpeven, zxiao10, jbai12, yzhang286, wqiu7, ayuille1, hager\}@jhu.edu

}

\usepackage{graphicx}
\usepackage{tabularx}
\usepackage{comment}
\usepackage{amsmath,amssymb} 
\usepackage{color}
\usepackage{epsfig}
\usepackage{paralist}
\usepackage{subcaption}
\usepackage{enumitem}
\usepackage{wrapfig}
\usepackage{float} 
\usepackage{everyshi}
\usepackage{dblfloatfix}

\usepackage{amsmath}
\usepackage{amssymb}
\usepackage{amsthm}
\usepackage{mathtools}

\DeclareMathOperator*{\argmax}{arg\,max}

\usepackage{tikz}
\usetikzlibrary{automata,fit,positioning,arrows}

\begin{document}

\maketitle

\begin{abstract}
There are many realistic applications of activity recognition where the set of potential activity descriptions is combinatorially large. This makes end-to-end supervised training of a recognition system impractical as no training set is practically able to encompass the entire label set. In this paper, we present an approach to fine-grained recognition that models activities as compositions of \textit{dynamic action signatures}.  This compositional approach allows us to reframe fine-grained recognition as \textit{zero-shot} activity recognition,  where a detector is composed ``on the fly'' from simple first-principles state machines supported by deep-learned components.  We evaluate our method on the Olympic Sports and UCF101 datasets, where our model establishes a new state of the art under multiple experimental paradigms. We also extend this method to form a unique framework for zero-shot joint segmentation and classification of activities in video and demonstrate the first results in zero-shot decoding of complex action sequences on a widely-used surgical dataset. Lastly, we show that we can use  off-the-shelf object detectors to recognize activities in completely de-novo settings with no additional training.
\end{abstract}

\section{Introduction} \label{sec:intro}
Fine-grained activity recognition is a challenging problem due to the combinatorial number of different fine-grained activities that are possible, and the variety of ways that they may be performed. For example, in video surveillance, we are most often interested in unique instantiations of an event — for example “locate an instance where a man wearing a yellow jacket approaches a blue SUV and places a brown package underneath it.”  This particular combination of actor, attributes, activity, and sequence is unique, and represents only one of millions of other activity sequences that are similar in structure, but unique in detail.

In this paper, we introduce the notion of \textit{dynamic action signatures} as a representation for fine-grained problems where there are a combinatorial number of potential fine-grained activity labels.  The key observation is that we can translate this problem into one of zero-shot activity recognition, where a recognition method is constructed “on the fly" using a composition of structured and deep-learned components.  As we show in our experiments, this method is highly flexible, and can be adapted to problem domains ranging from video surveillance to surgery.

\begin{figure*}[t]
\centering
\includegraphics[width=1.0\linewidth]{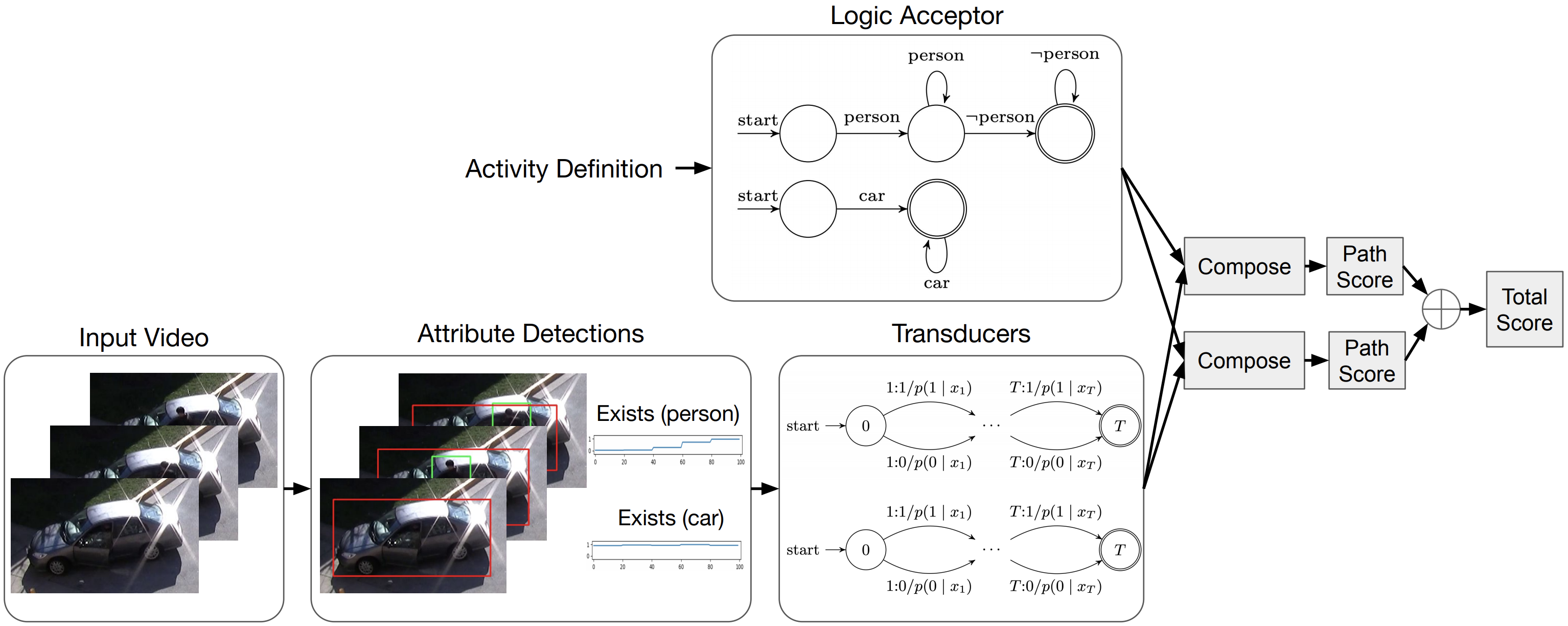}
  \caption{Our full system diagram. One path of the diagram creates transducers from neural attribute detections which get composed with the state-machine instantiated from the activity definitions. The best-path decode computes a compatibility score of a hypothesized activity.}
\label{fig:money}
\end{figure*}
In our framework, an action signature is a particular configuration of visually detectable entities, such as attributes, objects, and relations, which describe a segment of a video. A key element of our approach is that we view these signatures as time-varying, rather than static---\emph{i.e.} an action's attributes change over time in a characteristic manner. For example, the act of \textit{a person entering a vehicle} as shown in Figure \ref{fig:money} (Logic Acceptor)
can be described as the attribute sequences \texttt{a person exists} followed by \texttt{a person does not exist} composed with a \texttt{a vehicle exists} relational constraint.
 
In the remainder of this paper, we show that dynamic action signatures provide a powerful semantic label embedding for zero-shot action classification and establish a new state-of-the-art on the Olympic Sports \cite{olympic-sports-2010} and UCF101 \cite{ucf-101} datasets.
We also use our methodology to impose constraints \emph{on the predicted action sequences themselves}, leading to the first zero-shot segmentation results on complex action sequences in a widely-used surgical dataset \cite{jigsaws-2014}, and establish, for the first time, a zero-shot baseline result that is competitive with end-to-end trained methods. 
 
The aforementioned experiments required training using samples from the target dataset to acquire action signature detectors. We show that we can eliminate any kind of supervised training on the dataset from which unseen (test) cases are drawn by using publicly available, off-the-shelf object detectors to construct action signatures. We combine this with our framework to provide a true \emph{de novo} model of an activity.  We evaluate quantitatively and qualitatively our zero-shot framework using these ``on the fly'' models on the challenging DIVA dataset\footnote{https://actev.nist.gov/}, which contains fine-grained human-object interactions under a real world video security footage.

\noindent In summary, the main contributions of the paper are:
\begin{itemize}
\item A zero-shot classification of actions with dynamic action signatures which establishes a new state-of-the-art on Olympic Sports \cite{olympic-sports-2010} and UCF101 \cite{ucf-101} datasets. We outperform all other methods for the standard zero-shot evaluation regardless of training assumptions (inductive/transductive).
\item To the best of our knowledge, we are the first to demonstrate zero-shot decoding of complex action sequences. We present our  results on a surgical dataset, JIGSAWS \cite{jigsaws-2014}, to jointly segment and classify fine-grained surgical gestures where we establish a strong baseline.
\item A demonstration of zero-shot classification of fine-grained human-object interactions in security footage that requires no supervised training of attribute detectors by leveraging off-the-shelf object detectors. 
\end{itemize}



\section{Related Work} \label{sec:related_work}
\textbf{Rule-based approaches}: Zero-shot classifiers require descriptions of novel activities provided either by humans or existing knowledge bases. In the seminal approach of \cite{Ivanov2000RecognitionOV}, descriptions of activities come from an attribute based stochastic context-free grammar which encodes expert knowledge to recognize activities. Variants of probabilistic graphical models such as Markov Logic networks \cite{markov_logic}, And-Or graphs \cite{attributeGrammar} and attribute-multiset grammars \cite{amg} have been proposed since then to implement temporal rules to provide stronger representation of activities. Such rule-based approaches effectively modeled temporal behavior of activities in a true zero-shot manner using a set of selected attributes. Our approach is similar in that temporal rules and detections are parsed using structured state machines to score observed sequences. However, the biggest criticism of this line of work was its inability to scale due to the hand-crafted nature of attribute set-selection and rule-design.  We claim in this paper and support with our experiments that a small set of (four) temporal `rules' implemented using our dynamic action signatures are scalable and not problem specific. We empirically show that they scale effectively across datasets and even across application domains.
Moreover, we show evidence that zero-shot methods can move-away from ad-hoc selection of attributes by demonstrating how our framework utilizes generic features such as off-the-shelf object detectors to compose zero-shot action classifiers.


\textbf{Visual attributes as semantic descriptions of activity labels:}
Compared to rule-based approaches from the past, recent zero-shot methods focus on mapping a deep feature representation of a video and some semantic description of a label into a common embedding space where similarity functions or simple classifiers can be learned.
 The line of work that uses attribute based semantic embedding for zero-shot action classification followed a pioneering work that originally proposed to categorize novel objects using visual attributes \cite{Lampert_IAP_DAP_cvpr_09}. As a natural extension, manually defined visual attributes are widely used to provide semantic descriptions of human actions for zero-shot learning \cite{Akata_SJE_CVPR2015,Fu2014,Gan_2015,Lampert_IAP_DAP_cvpr_09,liu-kuipers-savarase-2011,Wang2017}. Using a fixed collection of attributes, a given sample is embedded as a vector of binary \cite{Lampert_IAP_DAP_cvpr_09,liu-kuipers-savarase-2011} or soft assignments \cite{Fu_2014,Romera_embarrassingly_simple_icml2015} of an attribute in the input. These attributes define a powerful semantic embedding space, as evidenced by recent approaches \cite{Gao_TSGCN_aaai2019,Mandal_OD_2019_CVPR,Mishra_GA_wacv_2018,Qin_ZSECOC_2017_CVPR, Brattoli_2020_CVPR} that consistently outperform word-embedding based approaches on zero-shot human action classification benchmarks \cite{olympic-sports-2010,ucf-101}. Such methods perceive activities as a collection `static' action signatures, limited to either presence or absence of a particular set of attributes in a given action sequence. In this work, we present \textit{dynamic} action signatures where an attribute can exhibit state changes over time.

\textbf{Word embedding as semantic descriptions of activity labels}: Paired with improvements in natural language processing and parsing \cite{mikolov_2013}, the popular Word2vec implementation \cite{word2vec} has been used for word embeddings in zero-shot systems \cite{Akata_SJE_CVPR2015,Alexiou2016ExploringSA,Gao_TSGCN_aaai2019,Guadarrama2013,Liu2019,Mandal_OD_2019_CVPR,Hahn2019Action2VecAC,Mishra_GA_wacv_2018,Qin_ZSECOC_2017_CVPR,democracy,Wang2017,Huang:2018:LJM:3240508.3240614,Zhang2018CrossModalAH,Zhang_glove_2018VisualDS,xu-hospedales-gong-2017,Xu2016}. The fundamental assumption of word embedding based approaches is that an activation pattern in the feature layer of the skip-gram model \cite{word2vec} given an interest-word (text of activity label) as input provides a discriminative representation of novel categories. 
Despite offering a clean manual-labor free solution compared to the attribute-based embeddings, zero-shot approaches using manual attributes consistently outperform Word2vec based models given the same methodology \cite{survey2019}. We believe an important factor in zero-shot action classification (as opposed to the static application of image classification) is to model the temporal evolution of elements in video. A text embedding of key words in its current form does not fully capture the dynamic structure present in activities.

\textbf{Objects as attributes:} The work of \cite{Jain_objects2action} constructs a word embedding augmented by a skip-gram model of object categories in videos. Further, a spatial-aware object based embedding is proposed in \cite{MettesICCV2017} for additional zero-shot localization of actions. An approach to learn relations between action-attribute-object in an end-to-end manner using two-stream graph convolutional network is proposed in \cite{Gao_TSGCN_aaai2019}. We also view objects as a promising source of additional information to provide a rich semantic embedding for zero-shot activity recognition. Our approach allows temporal modeling of object level information and we demonstrate that we can achieve zero-shot recognition of actor-object interactions using off-the-shelf object detectors in our experiments using the DIVA dataset.

\textbf{Domain Shift:} In practice, the distributions of features from seen and unseen categories are often not well aligned for accurate zero-shot inference especially when using pretrained deep features. Researchers have identified this problem as the domain shift problem, analyzed empirically in \cite{Fu_2014} and theoretically in \cite{Romera_embarrassingly_simple_icml2015}. This has led to a series of approaches \cite{Fu2014,xu-hospedales-gong-2017,uda_2015,Mishra_GA_wacv_2018,Gao_TSGCN_aaai2019,Mandal_OD_2019_CVPR} that allow the use of unlabeled instances from the unseen test categories as part of training, defined as the transductive setting for zero-shot learning. In this work, we focus on the introduction of a novel semantic embedding space with dynamic action signatures to improve temporal representation of activities, rather than a method to alleviate the domain shift between seen and unseen categories.  We note that the recent work of \cite{Brattoli_2020_CVPR} aims to completely move away from training on videos sampled from the target dataset by pretraining an end-to-end video-to-word-embedding network using large-scale independent video datasets. We also believe a more practical zero-shot system should be able to utilize available resources such as off-the-shelf word embeddings or object detectors to recognize true \textit{de novo} instances without having to finetue on samples from the target dataset. Our DIVA experiments demonstrate how our zero-shot system can be deployed in this open-set setting.



\section{Methodology}
\label{sec:methodology}
We first formulate the zero-shot action classification and describe the \textit{static} action signature in Section \ref{sec:static}.  In Section \ref{sec:activity_signatures}, we then introduce \textit{dynamic} action signatures for zero-shot problem and highlight the differences with the static counterpart. Then, in Section \ref{sec:fst}, we include a brief overview of finite state machines which are augmented to implement dynamic action signatures as described in Section \ref{sec:das_fst}. Finally, in Section \ref{sec:complex_activities}, we introduce how the same framework can be extended to model complex activities as sequences of actions to solve zero-shot temporal activity segmentation applications.


\begin{figure*}[t]
\centering
\includegraphics[width=\linewidth]{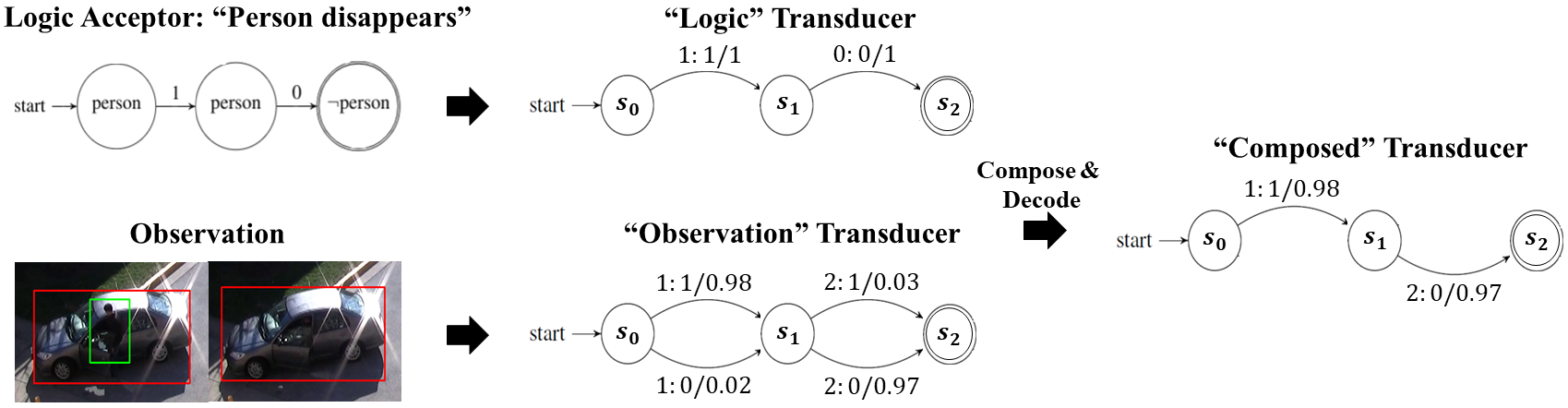}
  \caption{The figure illustrates how our framework scores an example observation with two frames. The `Logic' acceptor describes the dynamic signature of ``present, then absent". The `Observation' transducer produces a time series of detection probabilities. The sum of paths in the composed machine can be viewed as a compatibility score of the logic and the observation. In this particular example, the sum of weights in the `Composed` transducer is large which suggests that our observation is well-aligned with `person disappears' dynamic action signature. Please refer to Sections \ref{sec:fst} and \ref{sec:das_fst} for notation definitions.} 
\label{fig:signature-sm}
\end{figure*}

\label{sec:fst_overview}
\subsection{Zero-shot Action Classification and Static Action Signatures}
\label{sec:static}
The goal of the zero-shot action classification task is
to discriminate between a set of action categories
$\mathcal{Y} = \{ y_1, \ldots, y_N \}$
that the classifier has never encountered in its training data.
It is assumed in this setting that although we have
no example videos from the unseen categories,
the categories themselves are known to the user.
Thus,  zero-shot action recognition methods require
mapping both a video $x = (x_1, \ldots, x_T)$ and a label $y$
to a shared feature space---specifically,
one in which a video's representation is closest to
that of its ground-truth label under some similarity function $s$:
\begin{equation}
    y_{gt} = \argmax_{y \in \mathcal{Y}} s(f(y), g(x))
\end{equation}
In general this space can be abstract. For example,
the video's mapping function $g$ can be derived from visual features,
and the label's mapping function $f$ can be derived from a word vector representation.

\emph{Attribute-based} zero-shot methods define this space so it corresponds to a set of semantic concepts $\{ a_1, \ldots, a_K \}$.
In this case the label's representation is a $K$-dimensional ``action signature" $a(y)$,
whose elements are set to one if the attribute is present or zero if it is not.
Likewise, the video's representation is a set of attribute detections $\hat{a}(x) \in \mathbb{R}^K$.
\subsection{Dynamic Action Signatures}
\label{sec:activity_signatures}
Previous work has defined the label's action signature as a single vector,
and reduced either the video representation or the similarity function along the time dimension.
However, in many scenarios the actions of interest are distinguished by their time evolution
rather than the presence or absence of static signatures.
Take ``person entering a vehicle" and ``person exiting a vehicle", for example.
Both of these actions share the static action signature \texttt{vehicle present} and \texttt{person present}.
However, they are differentiated from each other by what happens to the person over time---%
in an \texttt{entering} action the person disappears into the vehicle,
but in an \texttt{exiting} action the person appears out of it.

In this paper, we make a very natural extension to this approach by extending an action signature in the temporal dimension.
This gives us a \textbf{dynamic action signature} of the form $a(y) = (a(y, 1), \ldots, a(y, T))$ that may change over time.
In the case that the value of an action signature is specified for every sample $t$,
the best category $y^*$ can be chosen using a straightforward sample-wise comparison:
\begin{equation} \label{eq:das-simple}
    y^* = \argmax_{y \in \mathcal{Y}} s(\hat{a}(x), a(y))
        = \argmax_{y \in \mathcal{Y}} \sum_{t=1}^{T} \sigma (\hat{a}(x_t), a(y, t))
\end{equation}
%
In Eq. \ref{eq:das-simple}, $\sigma(\cdot, \cdot)$ implements a sample-level similarity score---%
for example, the inner product.

Our framework can also handle the more interesting situation in which only the \emph{temporal ordering} of attributes is specified---%
for example, for the \texttt{entering vehicle} category we expect to see \texttt{person detected} $\rightarrow$ \texttt{person not detected},
but we likely do not know exactly \emph{when} the transition occurs.
In this case there can be many attribute sequences that are consistent with the specified ordering,
and we can define the similarity between a video $x$ and a category $y$
as the best value over that set (which we will call $\mathcal{A}$):
\begin{equation} \label{eq:fst-score}
    s(\hat{a}(x), a(y)) = 
        \argmax_{a(y) \in \mathcal{A}}
            \sum_{t=1}^{T} \sigma (\hat{a}(x_t), a(y, t))
\end{equation}
One of the key ideas in this paper is that we can describe the set $\mathcal{A}$ using a composition of a small vocabulary of finite state machines, and we can frame recognition as finding the maximum weight path in that machine.  We note that these machines can be either learned from data, or can be hand-specified. Here we pursue the latter approach; for clarity we review the fundamentals of finite state methods in the next section.



\subsection{Background: Finite State Machines}
\label{sec:fst}
%

A finite state acceptor (FSA) is a directed graph whose edges are labeled with input symbols.
Each vertex $s_t$ represents a state of the machine at time step $t$,
and each edge represents the state transition induced by observing a particular symbol $(i)$.
When the edges of an FSA carry a weight $w$, it is a weighted FSA (WFSA); in this case we write the edges as a pair $(i/w)$.
The total weight of any path in a WFSA is defined to be the product of the weights on each edge in the path.

An automaton that has been augmented to map input sequences to output sequences,
rather than only accept a set of inputs,
is called a weighted finite state transducer (WFST).
The edges of a WFST are labeled with a triple, $(i:j)/w$, defining the input symbol $i$,
the output symbol $j$, and the weight $w$ associated with that input-output pair. For example, the transducer shown in Figure \ref{fig:signature-sm} (``Observation Transducer'') maps sample indices (time steps) to attribute detections.

\textbf{Composition of WFSTs}: The `Logic' Transducer $T_1$ (see Figure \ref{fig:signature-sm}) provides a description of a particular dynamic pattern. Given an `Observation' transducer $T_2$, we can compute the compatibility between the pattern described by $T_1$ and the observation via composition of the two WFSTs.

Let $e_1(t) = (i_1:j_1)/w_1 \in T_1$ and $e_2(t) = (i_2:j_2)/w_2 \in T_2$ be edges in their respective transducers at a common time-step $t$ (both edges originate from state $s_t$ in the respective transducers).
For all such pairs of edges $e_1(t), e_2(t)$ where $j_1=i_2$, two edges can be composed as a single edge in a composed transducer $T=T_1\circ T_2$ with a composed edge $e(t) = (i_e:j_e)/w_e$ such that:
\begin{equation}
\label{eq:compose}
    i_e = i_1, j_e = j_2, w_e=w_1*w_2
\end{equation}
Please refer to the `Composed' transducer in Figure \ref{fig:signature-sm} as a concrete example.





When using finite state machines for sequence prediction tasks,
one is usually concerned with finding an optimal path
(eg the most probable output sequence given a particular input sequence),
or with finding the total weight of a set of paths
(eg the total probability of the data).
These quantities can be computed using generalizations of the well-known Viterbi
and Forward algorithms (respectively) for hidden Markov models/linear-chain CRFs.

\subsection{Dynamic Action Signatures as Finite State Machines}
\label{sec:das_fst}
%

We now return to Equation \ref{eq:fst-score} where the task is to find the dynamic action signature $a(y)$ in set $\mathcal{A}$ that produces the best score given attribute detections $\hat{a}_{1:T}$. The set $\mathcal{A}$ of all sequences that are consistent with the rule
can be described by a left-to-right weighted finite state acceptor $A$, the `Logic' Acceptor, with edges of the form $e_a=(i/w_a)$ where $i= \{0,1\}$ and $w_a=1$.
Figure \ref{fig:signature-sm} shows an example `Logic' Acceptor for an input with only two time-steps which accepts one `1` followed by one `0'. Intuitively, the machine generally describes a sequence where an attribute signal is present early and then is absent later in the sequence (\texttt{person detected} $\rightarrow$ \texttt{person not detected}). The rule acceptor $A$ can be re-written as a WFST $T_a$, the `Logic' Transducer in Figure \ref{fig:signature-sm}, by replicating inputs and output such that an edge is defined as $e_a=(i:i)/w_a$ and $w_a=1$. 

Then, we instantiate a WFST $T_b$ that accepts sample indices $t$ as input,
gives detections (0 or 1) as output,
and whose weights $p(1\  \text{or}\  0|x_t)$ correspond to attribute detection scores
(the `Observation' Transducer in Figure \ref{fig:signature-sm}). In the running example, $T_b$ describes a time-series of probability of human presence. 
Finally, given $T_a$ and $T_b$, we obtain the `Composed' Transducer $T=T_b \circ T_a$ as defined in Equation \ref{eq:compose}. Then, the solution to Equation \ref{eq:fst-score} is equivalent to finding the most probable `Logic' acceptor, $a(y)$, which yields the highest product of path weights when composed with the observation (we take the log sum instead in practice).


\subsection{Complex Activities as a Sequence of Actions}
\label{sec:complex_activities}
When a video is labeled with a \emph{sequence} of actions $y = (y_1, \ldots, y_M)$ (as opposed to sequence of attributes),
we can extend the zero-shot classification scenario to perform joint classification and segmentation in a zero-shot manner by defining a sequence-level score function over $M$ hypothesized segments:

\begin{equation}
\label{eq:segmental-score}
    y^* = \argmax_{y \in \mathcal{Y}^{\otimes T}}
        \sum_{i = 1}^{M}
            s(\hat{a}(x_{t_i:t_i + d_i}), a(y_{t_i:t_i + d_i}))
\end{equation}

In equation \ref{eq:segmental-score},
the quantity inside the sum implements the segment-level score function of Eq. \ref{eq:fst-score}
(or as a special case, Eq. \ref{eq:das-simple}),
defined over the $i$-th hypothesized segment with start time $t_i$, duration $d_i$, and label $y_i$.

In many cases, these activities have a structure that is known \emph{a priori}, and which can be exploited to rule out impossible action sequences.
For example, the JIGSAWS dataset is composed of surgical suturing videos.
In these sequences, only certain gesture sequences are realizable.
We can restrict the system's search space to valid sequences 
rather than the set of all possible action sequences $\mathcal{Y}^{\otimes T}$.
The best segmentation can be computed using a dynamic program---%
for example, one of the algorithms presented in \cite{Colin_2016} or \cite{sarawagi-cohen-2004}.
\section{Experiments}

In the following sections, we discuss the datasets that we evaluate on and how our method can be applied in a zero-shot manner on datasets built for action classification \emph{and} datasets built for joint classification \& segmentation. 



\subsection{Datasets} \label{sec:datasets}

\subsubsection{Olympic Sports and UCF101: Zero-shot Action Classification} 
We evaluate the benefit of dynamic action signatures for zero-shot action classification using the Olympic Sports \cite{olympic-sports-2010} and UCF101 \cite{ucf-101} datasets. We adhere to the settings proposed by \cite{xu-hospedales-gong-2017} and perform 30 independent test runs with randomly chosen seen/unseen classes (8/8 for Olympics, 51/50 for UCF101) for both standard (ZSL) and generalized (GZSL) zero-shot evaluations. We report the mean-per-class accuracy over 10 trials along with the standard deviation over trials. As reported in \cite{Mandal_OD_2019_CVPR}, we report mean accuracies for both seen and unseen categories in the test set and compute the harmonic mean of the two.

\subsubsection{JIGSAWS}
Here, we demonstrate that we can extend our framework to model complex activities as sequences of actions. We use the JIGSAWS dataset \cite{jigsaws-2014} to evaluate our framework on a more complex task: joint segmentation and classification. This is a publicly available dataset containing 39 instances of eight surgeons performing a benchtop simulation training task for robotic surgery. The dataset includes endoscopic video of the performance with per-frame gesture class labels for 10 types of actions that occur during the task. JIGSAWS only provides annotations for gestures (activities), and so we use the method described in Section \ref{sec:dynamic_attributes} to obtain per-frame attribute annotations.

\subsubsection{DIVA}
The DIVA dataset (extended from the VIRAT \cite{oh2011large} dataset) consists of very long video sequences captured from 5 independent camera viewpoints. It is a challenging activity detection benchmark where strong end-to-end baselines such as \cite{Rc3d} performs very poorly.
Factors such as lack of sufficient training data and large intra-class variance across camera viewpoints lead to poor performance of deep network based approaches \cite{Gleason2018APS,Kim2019SAFERF}. However, \cite{Kim2019SAFERF} has shown that objects such as vehicles and humans can be detected reliably. With DIVA experiments, we demonstrate that off-the-shelf object detectors can be used to compose dynamic action signatures to perform zero-shot activity recognition and may even out-perform fully-supervised baselines under such data conditions.


\begin{table*}[t]

\centering
\begin{tabular}{|c|cc|cc|cc|}
\hline
 &  &  & \multicolumn{2}{c|}{ZSL}& \multicolumn{2}{c|}{GZSL} \\ \hline
Method & Emb & ID/TD & Olympic & UCF101 & Olympic & UCF101 \\ 
\hline
GA \cite{Mishra_GA_wacv_2018} & W & ID & $34.1 \pm 10.1$ & $17.3 \pm 1.1$ & - &-\\
DAP \cite{Lampert_IAP_DAP_cvpr_09} & A & ID & $45.4 \pm 12.8$ & $15.9 \pm 1.2$ & -& -\\
HAA  \cite{liu-kuipers-savarase-2011} & A & ID & $46.1 \pm 12.4$ & $14.9 \pm 0.8$ & $49.4 \pm 10.8$ & $18.7 \pm 2.4$ \\
SJE \cite{Akata_SJE_CVPR2015} & A & ID & $47.5 \pm 14.8$ & $12.0 \pm 1.2$ & $32.5 \pm 6.7$ & $8.9 \pm 2.2$\\
GA \cite{Mishra_GA_wacv_2018} & A & ID & $50.4 \pm 11.2$ & $22.7 \pm 1.2$ & - & -\\
GCN \cite{Gao_TSGCN_aaai2019} & W & ID & $56.5 \pm 6.6$  & $34.2 \pm 3.1$ & - & -\\
E2E+K664* \cite{Brattoli_2020_CVPR} & W & ID & -  & $48.0 \pm 0.0$ & - & -\\
\hline
\textbf{DASZL (Ours)} & A & ID & $\textbf{74.2} \pm \textbf{9.9}$ & $\textbf{48.9} \pm \textbf{5.8}$ & $\textbf{56.4} \pm \textbf{9.4}$ & $\textbf{43.8} \pm \textbf{3.5}$\\ \hline
\hline
OD \cite{Mandal_OD_2019_CVPR} & W & TD & $50.5 \pm 6.9$ & $26.9 \pm 2.8$ & $53.1 \pm 3.6$ & $37.3 \pm 2.1$\\
GA \cite{Mishra_GA_wacv_2018} & A & TD & $57.8 \pm 14.1$ & $24.4 \pm 2.9$ & $52.4 \pm 12.2$ & $23.7 \pm 1.2$\\
GCN \cite{Gao_TSGCN_aaai2019} & W & TD & $59.9 \pm 5.3$ & $41.6 \pm 3.7$ & $50.2 \pm 6.8$ & $33.4 \pm 3.4$\\
f-GAN \cite{Xian_feature_GAN_2018cvpr}& A & TD & $64.7 \pm 7.5$ & $37.5 \pm 3.1$ & $59.9 \pm 5.5 $ & $44.4 \pm 3.0$\\
OD \cite{Mandal_OD_2019_CVPR} & A & TD & $65.9 \pm 8.1$ & $38.3 \pm 3.0$ & $\textbf{66.2} \pm \textbf{6.3}$ & $\textbf{49.4} \pm \textbf{2.4}$\\ 
\hline
\textbf{DASZL (Ours)} & A & TD & $\textbf{74.2} \pm \textbf{9.9}$  & $\textbf{48.9} \pm \textbf{5.8}$ & $61.0 \pm 4.9$  & $45.3 \pm 2.8$\\ 
\hline
\end{tabular}

\caption{Comparison of methods on Olympic Sports and UCF101 datasets. Manually defined attributes (A), Word embeddings (W). E2E+K664 follows a different training protocol where 664 classes from Kinetics is used to pretrain the network.}

\label{tab:zsl_sota}

\end{table*}

\subsection{Implementation Details} \label{sec:training_and_inference}

To re-use attribute definitions already provided by public datasets, we first describe a simple approach to extend static signatures into dynamic ones. Next we provide information on how these dynamic attributes can be learned from video.

\subsubsection{Dynamic Attribute Annotations:} \label{sec:dynamic_attributes}
We highlight that both Olympic Sports and UCF101 already have publicly available binary attributes associated per action class. These annotations denote whether an attribute (e.g. ``Ball-like Object'') exists or not per action class. We extend these binary values by adding two temporal patterns. This totals to the following four dynamic attribute patterns, namely: $\texttt{(0):Absence}$, $\texttt{(1):Persistence}$, $\texttt{(2):Start}$ and \texttt{(3):End}. Please refer to the supplementary material for all dynamic action signatures defined for Olympic/UCF activities. For the JIGSAWS dataset, no attribute annotations exist but the total number of attributes and actions are small, leading to an efficient annotation effort. In realistic applications, the relevant attributes will be defined by the problem itself (eg. search where a `needle' is transferred from `left' to `right gripper') or will be given by an expert with surgical domain knowledge.


In DIVA, the required attribute set is described in the action definition itself. We re-emphasize that many realistic zero-shot applications such as the DIVA setting (as opposed to Olympics, UCF) are of this form. For example the action ``Person Enters Car'' already defines both ``Person'' and ``Car'' attributes. No annotations are necessary as off-the-shelf detectors are utilized.

\subsubsection{Attribute Detectors:}
For experiments on Olympic Sports, UCF101 and JIGSAWS, we finetune a pretrained TSM \cite{lin2019tsm} to predict attribute presence given a video segment. We sample a video snippet $x_{t:t+d}$ from a video $X$ where $d$ is the length of the snippet, $t+d < T$ and $T$ is the length of $X$. The presence of an attribute $a(x_{t:t+d})$ in the snippet is determined accordingly based on its dynamic attribute label (eg \texttt{(2):End}) and relative position of $t$ with respect to $T$. For example, given a video of length $T=100$ and a dynamic attribute label of \texttt{(2):End}, $a(x_{10:42})=0$ whereas $a(x_{70:90})=1$ for the two snippets extracted from $X$. Optimization settings are provided in the supplementary material.

\section{Results \& Discussion}

In Section \ref{sec:results_classification} we provide evaluations on the Olympic Sports \cite{olympic-sports-2010} and UCF101 \cite{ucf-101} datasets. We demonstrate that dynamic action signatures significantly improve zero-shot action classification performance. In Section \ref{sec:results_jigsaws}, we then broaden the scope and describe our approach for the joint segmentation and classification of activities on the JIGSAWS \cite{jigsaws-2014} dataset. Lastly, we demonstrate that off-the-shelf detectors can be used for zero-shot activity classification of fine-grained human-object interactions in the DIVA dataset using our framework.


\subsection{Zero-shot Action Classification} \label{sec:results_classification}

\subsubsection{Zero-shot training and testing settings}

In the zero-shot action recognition literature, there are two settings for evaluation: zero-shot learning (ZSL) and generalized zero-shot learning (GZSL). Additionally, there are two settings for training models: inductive (ID) and transductive (TD). We explain these here.

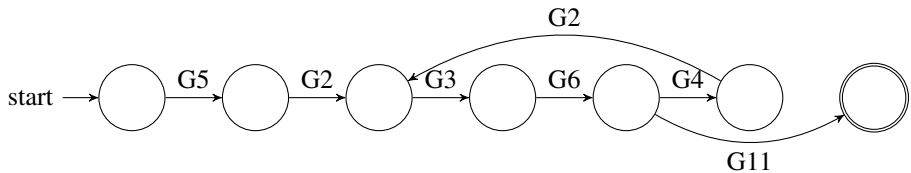
\begin{figure*}[!b]
    \centering
    \begin{tikzpicture}[->,node distance=0.75cm,>=stealth']
        \node[state, initial, inner sep=0pt] (g1) {};
        \node[state, right=of g1] (g5) {};
        \node[state, right=of g5] (g2) {};
        \node[state, right=of g2] (g3) {};
        \node[state, right=of g3] (g6) {};
        \node[state, right=of g6] (g4) {};
        \node[state, accepting, right=of g4] (g11) {};
        
        \draw
            (g1) edge[above] node{G5} (g5)
            (g5) edge[above] node{G2} (g2)
            (g2) edge[above] node{G3} (g3)
            (g3) edge[above] node{G6} (g6)
            (g6) edge[above] node{G4} (g4)
            (g6) edge[bend right, below] node{G11} (g11)
            (g4) edge[bend right, above] node{G2} (g2)
            ;
    \end{tikzpicture}
    \caption{
        State machine depicting an ideal suturing gesture sequence.
    }
    \label{fig:jigsaws-sm}
\end{figure*}

\textbf{Testing setup (ZSL/GZSL)} In the ZSL evaluation, the set of activities in the training videos (labelled videos) are disjoint from the set of activities in the test videos (unlabelled videos). Because test activities never overlap with training activities in the ZSL setting, it is not possible to evaluate whether models are biased to more accurately predict the activities they were trained on.
The GZSL evaluation was built to test the effect of this bias, and differs from the ZSL setting in that the test videos are drawn from the union of both the labelled and unlabelled activity sets.

\textbf{Training setup (ID/TD)} In the inductive (ID) setting, the training videos are from the labeled set of activities. In the transductive (TD) setting, both the labelled training videos and the unlabelled test videos are included during training. This training setup was built to mitigate the problem of domain-shift between the activity sets, and improve performance in the GZSL setting.
Our approach is inductive.

\subsubsection{State-of-the-art Comparisons}
In Table \ref{tab:zsl_sota}, the top half of the table compares models with the same training data assumptions (ID) as our approach under both evaluations. We highlight that under both tasks, our approach outperforms other ID models by a large margin. The ZSL evaluation provides a clean measure to directly compare the expressiveness of zero-shot systems for action classification because a classifier's bias to seen categories \cite{Mandal_OD_2019_CVPR} does not affect classification performance. We empirically show that by modeling actions using dynamic action signatures, we obtain a much more powerful zero-shot action classifier.

The bottom half of Table \ref{tab:zsl_sota} compares TD zero-shot methods. Again, our approach establishes a new state of the art under the ZSL evaluation even when compared to models with relaxed training data assumptions compared to our approach. This is a strong evidence that exploiting the temporal structure of actions leads to more gains in zero-shot classification performance than observing unlabeled test instances.
 
For the GZSL evaluation, with the ID version of our model, we report accuracies of $76.2 \%$ for seen and $44.7 \%$ for unseen categories (harmonic mean of $56.4 \%$) on Olympics. And consistently in UCF101, we see $67.2 \%$ on seen and $32.5 \%$ on unseen activities (harmonic mean of $43.8 \%$). We observe the domain shift problem mentioned in the previous section. However, we can adopt methods such as the binary in-vs-out-of-distribution auxiliary classifier ($OD_{aux}$) proposed in \cite{Mandal_OD_2019_CVPR} to implement a naive transductive solution. At inference time, the auxiliary binary prediction simply serves to limit the allowable prediction of our model which ultimately increases the recall of unseen categories at the cost of seen categories. For GZSL, the addition of $OD_{aux}$ reduces the performance gap between the seen and unseen categories which ultimately leads to gains of 4.6 and 1.5 on Olympics and UCF101. Since it is unnecessary to invoke the $OD_{aux}$ for the ZSL evaluation, we report identical results for ID/TD settings. We view a transductive extension of our framework as an interesting direction for future work.



\subsection{JIGSAWS: Zero-shot Action Classification \& Segmentation} \label{sec:results_jigsaws}
We have established that modeling actions as sequences of dynamic attributes leads to state of the art zero-shot classification performance. However, current zero-shot evaluation framework focuses on comparing models that all assume common semantic embedding of labels such as predefined attribute sets or word vector embeddings. Though such evaluation protocol provides fair and straight-forward comparisons between zero-shot methods, it fails to paint a clear picture of how zero-shot methods can extend beyond action classification. In the following sections, we show that we can extend our zero-shot framework easily to model activities as sequences of actions and thus enabling the first application of temporal zero-shot activity segmentation.

\subsubsection{Classification}


We begin by performing an ablation experiment studying the effectiveness of our dynamic signatures on the JIGSAWS dataset for the \emph{classification} setting.
For the static-signature system, we map all dynamic signatures to their nearest static counterparts.
Specifically, we map signatures \texttt{(2):start} and \texttt{(3):end} to \texttt{(0):absence}.
Our static-attribute system's average accuracy is 58.9\%,
while our system with dynamic attributes performs at 69.7\%.
Allowing dynamic signatures disambiguates between these gestures that are otherwise indistinguishable using only static signatures.


\subsubsection{Classification \& Segmentation}
We next turn to the task of zero-shot decoding,
\emph{joint classification and segmentation}, of surgical activity.
For this task, a long video which contains multiple actions is given and the task is to specify the start, end and label of all actions within a complex activity.
This task can be performed in a naive way by doing zero-shot classification for individual samples or windows of samples,
but frequently practitioners are aware of additional structure that restricts which action sequences are realizable.
In this experiment, we compare the performance of a grammar derived from first-principles knowledge of surgical suturing tasks with an unstructured baseline.


More specifically,
our grammar describes an ideal execution of the suturing task
(see fig. \ref{fig:jigsaws-sm} for an illustration).
The practitioner begins by reaching for the needle (G1),
then moves to the work area (G5), then executes a suture (G2 - G6).
At this point they can either transfer the needle from the left to the right hand (G4) and perform another suture,
or drop the suture and end the activity (G11).
Note that not every sequence in the JIGSAWS dataset conforms to this model---there are a few rare states (G8, G9, and G10)
that correspond to errors made during the suturing process.
Since this work addresses zero-shot applications,
we focus on modelling reliable and structured correct cases instead of the more variable incorrect ones.

The performance of our zero-shot decoding system and the unstructured baseline are recorded in Table \ref{tab:jigsaws-decoding-comparison}.
Our structured model improves frame-level accuracy over the unstructured one by about 8\%.
However, the improvement in edit score is much more substantial. By providing information about the basic structure of the sequence derived from what we know about the underlying process, we obtain close to a 100\% relative improvement in edit score.

\setlength{\tabcolsep}{4pt}
\renewcommand{\arraystretch}{1}
\begin{table}[t]
\begin{center}
\begin{tabular}{lcc}
Method & Edit Score & Acc (\%)  \\ \hline
IDT (s)  & 8.5 & 53.9 \\
VGG (s) & 24.3 & 45.9 \\
\textbf{DASZL w/o grammar} (z)  & 32.7 & 48.5 \\
\textbf{DASZL} (z) & 61.7 & 56.6 \\
Seg-CNN (s) \cite{Colin_SegCNN} & 66.6 & 74.7 \\
ST-CNN (s) \cite{Colin_SegCNN}  & 68.0 & 77.7 \\
TCN (s) \cite{Colin_2016}  & 83.1 & 81.4 \\
\end{tabular}
\caption{
    Comparison of our zero-shot method with previous supervised methods for joint classification and segmentation on JIGSAWS. (s): Supervised and (z): zero-shot methods
}
\label{tab:jigsaws-decoding-comparison}
\end{center}

\end{table}

Table \ref{tab:jigsaws-decoding-comparison} compares our method with previous, \emph{fully-supervised} ones. All listed methods except ours (DASZL variants) are fully supervised methods.
The first three supervised models represent unstructured, neural baselines established in \cite{Colin_2016}.
Interestingly, we obtain better edit distance than all of them and better accuracy than two of the three \emph{without ever training on gesture-labeled data}. Furthermore, our edit distance comes close to that of the segmental spatiotemporal CNN of \cite{Colin_2016}---a fully-supervised model that also incorporates a grammar.

\begin{table}[!t]
\begin{tabular}{ccc|cc}
\multicolumn{1}{l}{} & \multicolumn{2}{c|}{Supervised} & \multicolumn{2}{c}{Zero-shot} \\ \cline{2-5} 
                      & R(2+1)D             & TSM             & E2E-K664    & DASZL (ours)    \\ \hline
Accuracy              & 57.1                & 63.7            & 50.0        & \textbf{85.0}  
\end{tabular}
\caption{Accuracies R(2+1)D and TSM baselines supervised using action labels and a zero-shot baseline E2E-K664 on the DIVA dataset. By using object detectors to compose dynamic action signatures, we show that our model generalizes better than fully supervised baselines and is more accurate than the state-of-the-art true zero-shot method.}
\label{tab:diva}

\end{table}

\subsection{DIVA: Zero-Shot Interaction Classification with Off-the-Shelf Detectors} \label{sec:results_diva}

Both experiments on zero-shot classification of human activities and zero-shot segmentation of surgical gestures involve a supervised training step to obtain attribute detectors using instances from seen categories. In this section, we demonstrate that publicly available, off-the-shelf object detectors can be used to compose a system to classify human-object interactions in a truly supervision-less zero-shot manner. We demonstrate how we encode first-principles temporal logic to define activities using state machines combined with off-the-shelf object detectors to tackle this realistic open-set scenario.

\begin{figure}[t]
    \centering
    \includegraphics[width=1.0\linewidth]{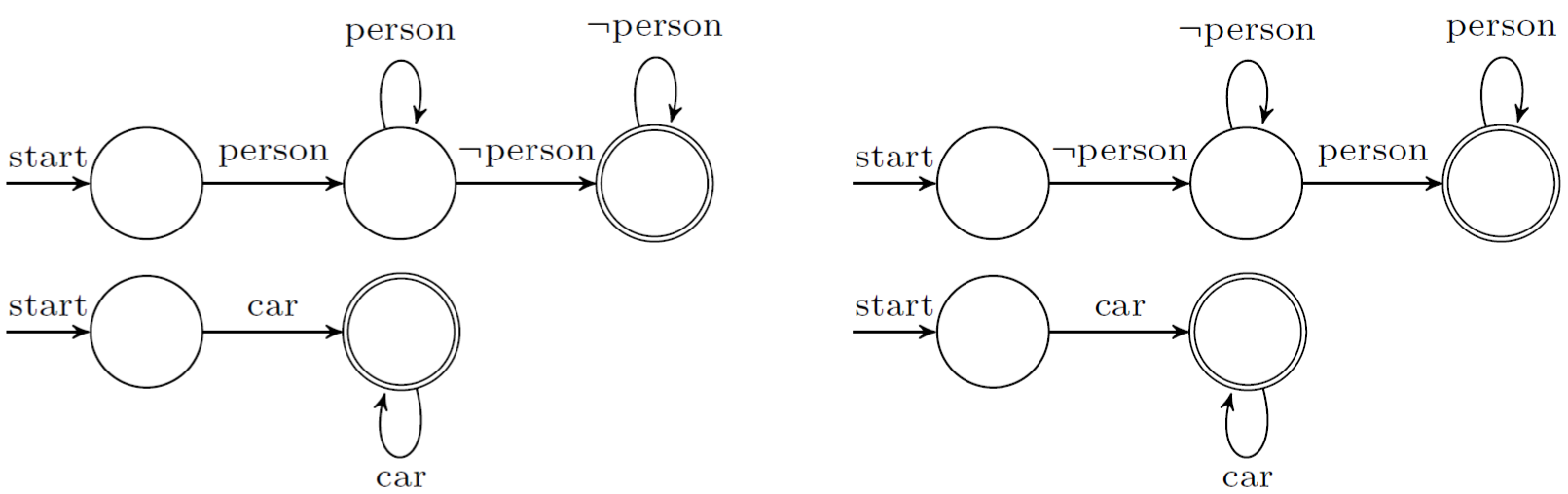}
    
    \caption{A set of state machines defined for \textit{Entering} (left) and \textit{Exiting} (right). A practitioner, using our framework, can effectively encode dynamic relations that may change over time to compose action classifiers in an open-set setting.}
    \vspace{-0.5cm}
     \label{fig:diva_states}
\end{figure}

Given detectable objects \{\texttt{Human}, \texttt{Vehicle}\}, we define a human \textit{Entering} and \textit{Exiting} a vehicle with state machines shown in Figure \ref{fig:diva_states}. We use publicly available object detectors \cite{He2017MaskR} and do not finetune the detectors. 

We compare our zero-shot system against well established end-to-end fully-supervised baselines, TSM  \cite{lin2019tsm}) and R(2+1)D \cite{res3d}. The DIVA dataset consists of instances sampled from five independent camera viewpoints where a model is trained on instances from four scenes and tested on samples from the held-out fifth scene (scene 0000). The results in Table \ref{tab:diva} show that our zero-shot approach outperforms the supervised baselines which are optimized end-to-end to predict activity labels given a video. This not only shows the discriminative power of dynamic action signatures for zero-shot methods but also emphasizes that the presented framework generalizes more robustly across viewpoints given its compositional nature. We also attempted to train TSM and R(2+1)D to jointly predict action labels and object presence but the models failed to converge.

We also compare against a truly zero-shot baseline \cite{Brattoli_2020_CVPR} which is trained to predict word embeddings directly given a video. We show that the E2E-K664 model performs poorly (chance is 50.0). We suspect that the word vector representations of \textit{Entering} and \textit{Exiting} are actually similar in the embedding space but for this classification task, we want the representations to be orthogonal. This is a fundamental limitation of activity representations derived from static word embeddings. The large performance gap corroborates that dynamic action signatures provide large benefits when modeling fine-grained interactions in a zero-shot manner.





\section{Conclusion}



We have presented a compositional approach for modeling fine-grained activities using dynamic action signatures. We showed that modeling actions as compositions spatio-temporal patterns of actors (attributes or objects) with temporal information improves zero-shot classification and we established a new state-of-the-art results on Olympic Sports and UCF101 datasets. Further, by modeling activities as sequences of actions, we established for the first a competitive baseline for a novel task of \textit{zero-shot segmentation} of complex surgical gesture sequences. Finally, with our framework, we showed an effective application of describing and recognizing fine-grained human-object interactions using first-principles knowledge combined with off-the-shelf object detectors.

 \textbf{Acknowledgements.} We gratefully acknowledge the support of NVIDIA Corporation with the donation of the Titan V GPUs used for this research. This work is supported by the Intelligence Advanced Research Projects Activity (IARPA) via Department of Interior/Interior Business Center (DOI/IBC) contract number D17PC00345. The U.S. Government is authorized to reproduce and distribute reprints for Governmental purposes notwithstanding any copyright annotation thereon. Disclaimer: The views and conclusions contained herein are those of the authors and should not be interpreted as necessarily representing the official policies or endorsements, either expressed or implied, of IARPA, DOI/IBC, or the U.S. Government.

\bibliography{egbib}

\begin{thebibliography}{47}
\providecommand{\natexlab}[1]{#1}
\providecommand{\url}[1]{\texttt{#1}}
\providecommand{\urlprefix}{URL }
\expandafter\ifx\csname urlstyle\endcsname\relax
  \providecommand{\doi}[1]{doi:\discretionary{}{}{}#1}\else
  \providecommand{\doi}{doi:\discretionary{}{}{}\begingroup
  \urlstyle{rm}\Url}\fi

\bibitem[{Akata et~al.(2015)Akata, Reed, Walter, Lee, and
  Schiele}]{Akata_SJE_CVPR2015}
Akata, Z.; Reed, S.~E.; Walter, D.; Lee, H.; and Schiele, B. 2015.
\newblock Evaluation of output embeddings for fine-grained image
  classification.
\newblock \emph{2015 IEEE Conference on Computer Vision and Pattern Recognition
  (CVPR)} 2927--2936.

\bibitem[{Alexiou, Xiang, and Gong(2016)}]{Alexiou2016ExploringSA}
Alexiou, I.; Xiang, T.; and Gong, S. 2016.
\newblock Exploring synonyms as context in zero-shot action recognition.
\newblock \emph{2016 IEEE International Conference on Image Processing (ICIP)}
  4190--4194.

\bibitem[{Brattoli et~al.(2020)Brattoli, Tighe, Zhdanov, Perona, and
  Chalupka}]{Brattoli_2020_CVPR}
Brattoli, B.; Tighe, J.; Zhdanov, F.; Perona, P.; and Chalupka, K. 2020.
\newblock Rethinking Zero-Shot Video Classification: End-to-End Training for
  Realistic Applications.
\newblock In \emph{Proceedings of the IEEE/CVF Conference on Computer Vision
  and Pattern Recognition (CVPR)}.

\bibitem[{Damen and Hogg(2009)}]{amg}
Damen, D.; and Hogg, D. 2009.
\newblock Attribute Multiset Grammars for Global Explanations of Activities.
\newblock In \emph{British Machine Vision Conference, BMVC 2009 - Proceedings}.

\bibitem[{Fu et~al.(2014)Fu, Hospedales, Xiang, Fu, and Gong}]{Fu2014}
Fu, Y.; Hospedales, T.~M.; Xiang, T.; Fu, Z.; and Gong, S. 2014.
\newblock Transductive Multi-view Embedding for Zero-Shot Recognition and
  Annotation.
\newblock In Fleet, D.; Pajdla, T.; Schiele, B.; and Tuytelaars, T., eds.,
  \emph{Computer Vision -- ECCV 2014}, 584--599.

\bibitem[{{Fu} et~al.(2014){Fu}, {Hospedales}, {Xiang}, and {Gong}}]{Fu_2014}
{Fu}, Y.; {Hospedales}, T.~M.; {Xiang}, T.; and {Gong}, S. 2014.
\newblock Learning Multimodal Latent Attributes.
\newblock \emph{IEEE Transactions on Pattern Analysis and Machine Intelligence}
  36(2): 303--316.
\newblock \doi{10.1109/TPAMI.2013.128}.

\bibitem[{Gan et~al.(2015)Gan, Lin, Yang, Zhuang, and Hauptmann}]{Gan_2015}
Gan, C.; Lin, M.; Yang, Y.; Zhuang, Y.; and Hauptmann, A.~G. 2015.
\newblock Exploring Semantic Inter-class Relationships (SIR) for Zero-shot
  Action Recognition.
\newblock In \emph{Proceedings of the Twenty-Ninth AAAI Conference on
  Artificial Intelligence}, AAAI'15, 3769--3775.

\bibitem[{Gao, Zhang, and Xu(2019)}]{Gao_TSGCN_aaai2019}
Gao, J.; Zhang, T.; and Xu, C. 2019.
\newblock I Know the Relationships: Zero-Shot Action Recognition via Two-Stream
  Graph Convolutional Networks and Knowledge Graphs.
\newblock In \emph{AAAI}.

\bibitem[{Gao et~al.(2014)Gao, Vedula, Reiley, Ahmidi, Varadarajan, Lin, Tao,
  Zappella, B{\'e}jar, Yuh, Chen, Vidal, Khudanpur, and Hager}]{jigsaws-2014}
Gao, Y.; Vedula, S.~S.; Reiley, C.~E.; Ahmidi, N.; Varadarajan, B.; Lin, H.~C.;
  Tao, L.; Zappella, L.; B{\'e}jar, B.; Yuh, D.~D.; Chen, C. C.~G.; Vidal, R.;
  Khudanpur, S.; and Hager, G.~D. 2014.
\newblock JHU-ISI Gesture and Skill Assessment Working Set (JIGSAWS): A
  Surgical Activity Dataset for Human Motion Modeling.
\newblock In \emph{Modeling and Monitoring of Computer Assisted Interventions
  (M2CAI) – MICCAI Workshop}.

\bibitem[{Gleason et~al.(2018)Gleason, Ranjan, Schwarcz, Castillo, Chen, and
  Chellappa}]{Gleason2018APS}
Gleason, J.; Ranjan, R.; Schwarcz, S.; Castillo, C.; Chen, J.-C.; and
  Chellappa, R. 2018.
\newblock A Proposal-Based Solution to Spatio-Temporal Action Detection in
  Untrimmed Videos.
\newblock \emph{2019 IEEE Winter Conference on Applications of Computer Vision
  (WACV)} 141--150.

\bibitem[{Guadarrama et~al.(2013)Guadarrama, Krishnamoorthy, Malkarnenkar,
  Venugopalan, Mooney, Darrell, and Saenko}]{Guadarrama2013}
Guadarrama, S.; Krishnamoorthy, N.; Malkarnenkar, G.; Venugopalan, S.; Mooney,
  R.; Darrell, T.; and Saenko, K. 2013.
\newblock YouTube2Text: Recognizing and Describing Arbitrary Activities Using
  Semantic Hierarchies and Zero-Shot Recognition.
\newblock In \emph{Proceedings of the 2013 IEEE International Conference on
  Computer Vision}, ICCV '13, 2712--2719. Washington, DC, USA: IEEE Computer
  Society.

\bibitem[{Hahn, Silva, and Rehg(2019)}]{Hahn2019Action2VecAC}
Hahn, M.; Silva, A.; and Rehg, J.~M. 2019.
\newblock Action2Vec: A Crossmodal Embedding Approach to Action Learning.
\newblock \emph{ArXiv} abs/1901.00484.

\bibitem[{He et~al.(2017)He, Gkioxari, Doll{\'a}r, and Girshick}]{He2017MaskR}
He, K.; Gkioxari, G.; Doll{\'a}r, P.; and Girshick, R.~B. 2017.
\newblock Mask R-CNN.
\newblock \emph{2017 IEEE International Conference on Computer Vision (ICCV)}
  2980--2988.

\bibitem[{Huang, Zhang, and Li(2018)}]{Huang:2018:LJM:3240508.3240614}
Huang, F.; Zhang, X.; and Li, Z. 2018.
\newblock Learning Joint Multimodal Representation with Adversarial Attention
  Networks.
\newblock In \emph{Proceedings of the 26th ACM International Conference on
  Multimedia}, MM '18, 1874--1882. New York, NY, USA.

\bibitem[{Ivanov and Bobick(2000)}]{Ivanov2000RecognitionOV}
Ivanov, Y.; and Bobick, A. 2000.
\newblock Recognition of Visual Activities and Interactions by Stochastic
  Parsing.
\newblock \emph{IEEE Trans. Pattern Anal. Mach. Intell.} 22: 852--872.

\bibitem[{Jain et~al.(2015)Jain, van Gemert, Mensink, and
  Snoek}]{Jain_objects2action}
Jain, M.; van Gemert, J.~C.; Mensink, T.; and Snoek, C. 2015.
\newblock Objects2action: Classifying and Localizing Actions without Any Video
  Example.
\newblock In \emph{ICCV}, 4588--4596. IEEE Computer Society.

\bibitem[{Kim et~al.(2019)Kim, Zhang, Xiao, Peven, Qiu, Bai, Yuille, and
  Hager}]{Kim2019SAFERF}
Kim, T.~S.; Zhang, Y.; Xiao, Z.; Peven, M.; Qiu, W.; Bai, J.; Yuille, A.~L.;
  and Hager, G.~D. 2019.
\newblock SAFER : Fine-grained Activity Detection by Compositional Hypothesis
  Testing.
\newblock In \emph{arxiv}.

\bibitem[{Kodirov et~al.(2015)Kodirov, Xiang, Fu, and Gong}]{uda_2015}
Kodirov, E.; Xiang, T.; Fu, Z.; and Gong, S. 2015.
\newblock Unsupervised Domain Adaptation for Zero-Shot Learning.
\newblock In \emph{ICCV}, 2452--2460.
\newblock \doi{10.1109/ICCV.2015.282}.

\bibitem[{Lampert, Nickisch, and Harmeling(2009)}]{Lampert_IAP_DAP_cvpr_09}
Lampert, C.~H.; Nickisch, H.; and Harmeling, S. 2009.
\newblock Learning to detect unseen object classes by betweenclass attribute
  transfer.
\newblock In \emph{In CVPR}.

\bibitem[{Lea et~al.(2017)Lea, Flynn, Vidal, Reiter, and Hager}]{Colin_2016}
Lea, C.; Flynn, M.; Vidal, R.; Reiter, A.; and Hager, G. 2017.
\newblock Temporal Convolutional Networks for Action Segmentation and
  Detection.
\newblock In \emph{cvpr}, 1003--1012.
\newblock \doi{10.1109/CVPR.2017.113}.

\bibitem[{Lea et~al.(2016)Lea, Reiter, Vidal, and Hager}]{Colin_SegCNN}
Lea, C.; Reiter, A.; Vidal, R.; and Hager, G.~D. 2016.
\newblock Segmental Spatiotemporal CNNs for Fine-Grained Action Segmentation.
\newblock In Leibe, B.; Matas, J.; Sebe, N.; and Welling, M., eds.,
  \emph{Computer Vision -- ECCV 2016}, 36--52.

\bibitem[{Lin, Gan, and Han(2019)}]{lin2019tsm}
Lin, J.; Gan, C.; and Han, S. 2019.
\newblock TSM: Temporal Shift Module for Efficient Video Understanding.
\newblock In \emph{Proceedings of the IEEE International Conference on Computer
  Vision}.

\bibitem[{Lin, Gong, and Li(2009)}]{attributeGrammar}
Lin, L.; Gong, H.; and Li, L. 2009.
\newblock Semantic event representation and recognition using syntactic
  attribute graph grammar.
\newblock \emph{Pattern Recognition Letters} 30: 180--186.

\bibitem[{{Liu}, {Kuipers}, and {Savarese}(2011)}]{liu-kuipers-savarase-2011}
{Liu}, J.; {Kuipers}, B.; and {Savarese}, S. 2011.
\newblock Recognizing human actions by attributes.
\newblock In \emph{CVPR 2011}, 3337--3344.
\newblock \doi{10.1109/CVPR.2011.5995353}.

\bibitem[{Liu et~al.(2019)Liu, Liu, Ma, Huang, and Dong}]{Liu2019}
Liu, K.; Liu, W.; Ma, H.; Huang, W.; and Dong, X. 2019.
\newblock Generalized zero-shot learning for action recognition with web-scale
  video data.
\newblock \emph{World Wide Web} 22(2): 807--824.

\bibitem[{Luis et~al.(2019)Luis, Junior, Pedrinib, and Menottic}]{survey2019}
Luis, V.; Junior, E.; Pedrinib, H.; and Menottic, D. 2019.
\newblock Zero-Shot Action Recognition in Videos: A Survey.
\newblock \emph{ArXiv} .

\bibitem[{Mandal et~al.(2019)Mandal, Narayan, Dwivedi, Gupta, Ahmed, Khan, and
  Shao}]{Mandal_OD_2019_CVPR}
Mandal, D.; Narayan, S.; Dwivedi, S.~K.; Gupta, V.; Ahmed, S.; Khan, F.~S.; and
  Shao, L. 2019.
\newblock Out-Of-Distribution Detection for Generalized Zero-Shot Action
  Recognition.
\newblock In \emph{The IEEE Conference on Computer Vision and Pattern
  Recognition (CVPR)}.

\bibitem[{Mettes and Snoek(2017)}]{MettesICCV2017}
Mettes, P. S.~M.; and Snoek, C. G.~M. 2017.
\newblock Spatial-Aware Object Embeddings for Zero-Shot Localization and
  Classification of Actions.
\newblock In \emph{IEEE International Conference on Computer Vision}.

\bibitem[{Mikolov et~al.(2013)Mikolov, Sutskever, Chen, Corrado, and
  Dean}]{word2vec}
Mikolov, T.; Sutskever, I.; Chen, K.; Corrado, G.; and Dean, J. 2013.
\newblock Distributed Representations of Words and Phrases and Their
  Compositionality.
\newblock In \emph{Proceedings of the 26th International Conference on Neural
  Information Processing Systems - Volume 2}, NIPS'13, 3111--3119. USA: Curran
  Associates Inc.

\bibitem[{Mikolov, Yih, and Zweig(2013)}]{mikolov_2013}
Mikolov, T.; Yih, W.-t.; and Zweig, G. 2013.
\newblock Linguistic Regularities in Continuous Space Word Representations.
\newblock In \emph{Proceedings of the 2013 Conference of the North {A}merican
  Chapter of the Association for Computational Linguistics: Human Language
  Technologies}, 746--751. Atlanta, Georgia: Association for Computational
  Linguistics.

\bibitem[{Mishra et~al.(2018)Mishra, Verma, Reddy, Subramaniam, Rai, and
  Mittal}]{Mishra_GA_wacv_2018}
Mishra, A.; Verma, V.~K.; Reddy, M. S.~K.; Subramaniam, A.; Rai, P.; and
  Mittal, A. 2018.
\newblock A Generative Approach to Zero-Shot and Few-Shot Action Recognition.
\newblock \emph{2018 IEEE Winter Conference on Applications of Computer Vision
  (WACV)} 372--380.

\bibitem[{Niebles, Chen, and Fei-Fei(2010)}]{olympic-sports-2010}
Niebles, J.~C.; Chen, C.-W.; and Fei-Fei, L. 2010.
\newblock Modeling Temporal Structure of Decomposable Motion Segments for
  Activity Classification.
\newblock In Daniilidis, K.; Maragos, P.; and Paragios, N., eds.,
  \emph{Computer Vision -- ECCV 2010}, 392--405. Springer Berlin Heidelberg.

\bibitem[{Oh et~al.(2011)Oh, Hoogs, Perera, Cuntoor, Chen, Lee, Mukherjee,
  Aggarwal, Lee, Davis et~al.}]{oh2011large}
Oh, S.; Hoogs, A.; Perera, A.; Cuntoor, N.; Chen, C.-C.; Lee, J.~T.; Mukherjee,
  S.; Aggarwal, J.; Lee, H.; Davis, L.; et~al. 2011.
\newblock A large-scale benchmark dataset for event recognition in surveillance
  video.
\newblock In \emph{CVPR 2011}, 3153--3160. IEEE.

\bibitem[{Qin et~al.(2017)Qin, Liu, Shao, Shen, Ni, Chen, and
  Wang}]{Qin_ZSECOC_2017_CVPR}
Qin, J.; Liu, L.; Shao, L.; Shen, F.; Ni, B.; Chen, J.; and Wang, Y. 2017.
\newblock Zero-Shot Action Recognition With Error-Correcting Output Codes.
\newblock In \emph{The IEEE Conference on Computer Vision and Pattern
  Recognition (CVPR)}.

\bibitem[{Roitberg, Al{-}Halah, and Stiefelhagen(2018)}]{democracy}
Roitberg, A.; Al{-}Halah, Z.; and Stiefelhagen, R. 2018.
\newblock Informed Democracy: Voting-based Novelty Detection for Action
  Recognition.
\newblock In \emph{British Machine Vision Conference 2018, {BMVC} 2018,
  Northumbria University, Newcastle, UK, September 3-6, 2018}, 52.

\bibitem[{Romera-Paredes and
  Torr(2015)}]{Romera_embarrassingly_simple_icml2015}
Romera-Paredes, B.; and Torr, P. H.~S. 2015.
\newblock An Embarrassingly Simple Approach to Zero-shot Learning.
\newblock In \emph{Proceedings of the 32Nd International Conference on
  International Conference on Machine Learning - Volume 37}, ICML'15,
  2152--2161. JMLR.org.

\bibitem[{Sarawagi and Cohen(2005)}]{sarawagi-cohen-2004}
Sarawagi, S.; and Cohen, W.~W. 2005.
\newblock Semi-Markov Conditional Random Fields for Information Extraction.
\newblock In Saul, L.~K.; Weiss, Y.; and Bottou, L., eds., \emph{Advances in
  Neural Information Processing Systems 17}, 1185--1192. MIT Press.

\bibitem[{Soomro et~al.(2012)Soomro, Zamir, Shah, Soomro, Zamir, and
  Shah}]{ucf-101}
Soomro, K.; Zamir, A.~R.; Shah, M.; Soomro, K.; Zamir, A.~R.; and Shah, M.
  2012.
\newblock UCF101: A dataset of 101 human actions classes from videos in the
  wild.
\newblock \emph{CoRR} 2012.

\bibitem[{{Tran} et~al.(2018){Tran}, {Wang}, {Torresani}, {Ray}, {LeCun}, and
  {Paluri}}]{res3d}
{Tran}, D.; {Wang}, H.; {Torresani}, L.; {Ray}, J.; {LeCun}, Y.; and {Paluri},
  M. 2018.
\newblock A Closer Look at Spatiotemporal Convolutions for Action Recognition.
\newblock In \emph{2018 IEEE/CVF Conference on Computer Vision and Pattern
  Recognition}, 6450--6459.

\bibitem[{Tran and Davis(2008)}]{markov_logic}
Tran, S.~D.; and Davis, L.~S. 2008.
\newblock Event Modeling and Recognition Using Markov Logic Networks.
\newblock In Forsyth, D.; Torr, P.; and Zisserman, A., eds., \emph{Computer
  Vision -- ECCV 2008}, 610--623.

\bibitem[{Wang and Chen(2017)}]{Wang2017}
Wang, Q.; and Chen, K. 2017.
\newblock Zero-Shot Visual Recognition via Bidirectional Latent Embedding.
\newblock \emph{International Journal of Computer Vision} 124(3): 356--383.

\bibitem[{{Xian} et~al.(2018){Xian}, {Lorenz}, {Schiele}, and
  {Akata}}]{Xian_feature_GAN_2018cvpr}
{Xian}, Y.; {Lorenz}, T.; {Schiele}, B.; and {Akata}, Z. 2018.
\newblock Feature Generating Networks for Zero-Shot Learning.
\newblock In \emph{2018 IEEE/CVF Conference on Computer Vision and Pattern
  Recognition}, 5542--5551.
\newblock ISSN 1063-6919.
\newblock \doi{10.1109/CVPR.2018.00581}.

\bibitem[{Xu, Das, and Saenko(2017)}]{Rc3d}
Xu, H.; Das, A.; and Saenko, K. 2017.
\newblock R-C3D: Region Convolutional 3D Network for Temporal Activity
  Detection.
\newblock In \emph{ICCV}, 5794--5803.
\newblock \doi{10.1109/ICCV.2017.617}.

\bibitem[{Xu, Hospedales, and Gong(2017)}]{xu-hospedales-gong-2017}
Xu, X.; Hospedales, T.; and Gong, S. 2017.
\newblock Transductive Zero-Shot Action Recognition by Word-Vector Embedding.
\newblock \emph{International Journal of Computer Vision} 123(3): 309--333.

\bibitem[{Xu, Hospedales, and Gong(2016)}]{Xu2016}
Xu, X.; Hospedales, T.~M.; and Gong, S. 2016.
\newblock Multi-Task Zero-Shot Action Recognition with Prioritised Data
  Augmentation.
\newblock In Leibe, B.; Matas, J.; Sebe, N.; and Welling, M., eds.,
  \emph{Computer Vision -- ECCV 2016}, 343--359.

\bibitem[{Zhang, Hu, and Sha(2018)}]{Zhang2018CrossModalAH}
Zhang, B.; Hu, H.; and Sha, F. 2018.
\newblock Cross-Modal and Hierarchical Modeling of Video and Text.
\newblock In \emph{ECCV}.

\bibitem[{Zhang and Peng(2018)}]{Zhang_glove_2018VisualDS}
Zhang, C.; and Peng, Y. 2018.
\newblock Visual Data Synthesis via GAN for Zero-Shot Video Classification.
\newblock In \emph{IJCAI}.

\end{thebibliography}

\end{document}